%% file: jrx_ukn_ali.tex
\documentclass[runningheads]{llncs}

\usepackage{graphicx}

\usepackage[inline]{enumitem}
\usepackage{amsmath, amssymb}
\usepackage{subcaption}
\usepackage{balance}
\usepackage{times}

\usepackage{subfiles}
\usepackage[misc]{ifsym}

\usepackage[colorinlistoftodos]{todonotes}
\usepackage{url}

\newcommand{\for}{\text{for }}

\newcommand{\True}{\mathit{True}}
\newcommand{\False}{\mathit{False}}
\newcommand{\Activated}{\alpha}
\newcommand{\Min}{\min}
\newcommand{\Max}{\max}
\newcommand{\Linked}{\mathit{CG}}
\newcommand{\NbPersons}{\omega}
\newcommand{\MaxNbPersons}{\Omega}
\newcommand{\CspEstimation}{\delta}

\begin{document}
\title{Online Guest Detection in a Smart Home using Pervasive Sensors and Probabilistic Reasoning\thanks{This work and the authors are supported by the distributed environment Ecare@Home (Swedish Knowledge Foundation 2015–2019) and the MoveCare project (European Commission H2020 framework program for research and innovation, ID 732158).}}

\author{Jennifer Renoux(\Letter)\and
  Uwe K\"{o}ckemann \and
  Amy Loutfi}
\authorrunning{J. Renoux et al.}
%
\institute{Center for Applied Autonomous Sensor Systems, \\
  \"{O}rebro University, Sweden \\
\email{firstname.lastname@oru.se}}

\maketitle

\begin{abstract}

Smart home environments equipped with distributed sensor networks are capable of helping people by providing services related to health, emergency detection or daily routine management. A backbone to these systems relies often on the system's ability to track and detect activities performed by the users in their home. Despite the continuous progress in the area of activity recognition in smart homes, many systems make a strong underlying assumption that the number of occupants in the home at any given moment of time is always known. Estimating the number of persons in a Smart Home at each time step remains a challenge nowadays. Indeed, unlike most (crowd) counting solution which are based on computer vision techniques, the sensors considered in a Smart Home are often very simple and do not offer individually a good overview of the situation. The data gathered needs therefore to be fused in order to infer useful information. This paper aims at addressing this challenge and presents a probabilistic approach able to estimate the number of persons in the environment at each time step. This approach works in two steps: first, an estimate of the number of persons present in the environment is done using a Constraint Satisfaction Problem solver, based on the topology of the sensor network and the sensor activation pattern at this time point. Then, a Hidden Markov Model refines this estimate by considering the uncertainty related to the sensors.  Using both simulated and real data, our method has been tested and validated on two smart homes of different sizes and configuration and demonstrates the ability to accurately estimate the number of inhabitants. %
 
\end{abstract}


\section{Introduction}
\subfile{sections/1-introduction-SotA.tex}

\section{The Model}
\label{sec:model}
\subfile{sections/2-model.tex}

\section{Evaluation}
\label{sec:evaluation}
\subfile{sections/3-evaluation.tex}

\section{Discussion}
\label{sec:discussion}
\subfile{sections/4-discussion.tex}

\section*{Acknowledgements}
This is a post-peer-review, pre-copyedit version of an article published in In European Conference on Ambient Intelligence (pp. 74-89). Springer, Cham. The final authenticated version is available online at: \url{https://doi.org/10.1007/978-3-030-03062-9_6}

\balance


\bibliographystyle{splncs04}
\bibliography{jrx_ukn_ali}

\end{document}

%% file: sections/1-introduction-SotA.tex
Ambient Assisted Living (AAL) technologies provide services such as health control, emergency detection systems or activity monitoring~\cite{wilson_smart_2015}.  Recent advancements in areas such as pervasive sensing, robotics and human-machine interaction bring these systems closer to real applications. Activity and context recognition are two very important aspects of these systems as they enable to act depending on the constantly evolving situation. In the case of multi-person activity recognition, group activity recognition has been attracting an important amount of work recently \cite{choi2012unified,ibrahim2016hierarchical}. However, these systems are computer vision based technique and are hardly usable in Smart Homes where the presence of cameras is not desirable. 
To our knowledge, all existing not vision based context and activity recognition techniques have been designed or trained for an assumed number of occupants in the home (usually one or two) and cannot handle other numbers of persons~\cite{benmansour_multioccupant_2016}.

However, real-life situations in our homes usually involve visits from other persons, especially in the domain of AAL where the users considered are often  living alone or with their partner, and are regularly visited by caregivers or guests. These unexpected visits, if undetected and unaccounted for, might degrade or even impede the possibility for such systems to make correct inference about alarms, activities of daily living and more. 

The ability to detect the presence of guests and  estimate the number of persons in the environment at a given time is of paramount importance for the development of AAL systems and their adaptation to real-life scenarios. The contribution of this work is a framework capable of estimating the number of occupants in a Smart Home using pervasive binary sensors. This framework combines a Constraint satisfaction Problem (CSP) Solver with a Hidden Markov Model (HMM) to handle uncertainty and recover from wrong estimations. 

The paper is presented as follows. First, we present a brief summary of existing work about person counting in various environments. Then Section \ref{sec:model} presents the developed model. The evaluation protocol and the results are presented in Section \ref{sec:evaluation}. Finally, Section \ref{sec:discussion} will discuss the capabilities and limitations of our approach. 

\section{Related Work}

While commercial solutions for person counting are said to exist, they use various sets sensors, such as thermal imagers or break beam sensors~\cite{teixeira_survey_2010} which suffer from a number of shortcomings. In addition to their costs, these solutions have the major disadvantage of not being robust to occlusion and are unable to recover from a false detection. They are therefore efficient only if one wishes to estimate a crowd or a stream of persons passing but, indeed of little use in a smart home where the counting needs to be more precise and robust. Other systems that allow people tracking and counting use vision-based sensors~\cite{prathiba_literature_2013,vera_counting_2016}. While these systems are efficient and reliable, the use of cameras makes them unsuitable for smart some environments for obvious privacy reasons. Finally, wearables such as RFID readers can be used to identify and track occupants of a Smart Home (e.g. in \cite{wang2011recognizing}, but they require the user to carry the device with them and are therefore not usable to detect persons who are only visiting and do not carry such a device. 

Pervasive sensing is becoming more efficient and affordable thanks to the rapid development of small low-cost sensors. This paradigm is already used in various of Ambient Intelligence applications~\cite{alirezaie_ontology_2017,noor_ontology_2018} and offers an ideal setup for an unobtrusive method of counting people in the environment. However, even though these applications require to know the number of occupants in the environment, they usually avoid this problem by assuming a fixed number of occupants and consequently, very little research is available where pervasive sensing is used for person counting. Some recent work focused on estimating the number of pedestrians in environments using binary sensors and Monte-Carlo methods~\cite{fujii_pedestrian_2014}, but these are again hardly usable in Smart Home situation due to various assumption about the topology of the environment. Recently,~\cite{renoux_context_2017} proposed a system using probabilistic reasoning to perform counting in Smart Home, but the assumptions made by the authors (total independence of the sensors) greatly limits its usability.

Counting the number of persons in a environment with unobtrusive environmental sensors under realistic hypotheses remains an open research question. In the next sections, we will be presenting our model aiming at addressing this issue while reducing the number of hypotheses considered.

%% file: sections/2-model.tex
We describe the input to our system as observations of \emph{Features of Interest (FoI)}.
A FoI for person counting is the representation of a real-world element that is monitored and indicates the presence of a person. 
A basic example of such a FoI is \emph{couch occupancy} which could be observed by a pressure sensor.
In the same way the FoI \emph{room presence} could be measured by a motion sensor.
More complex FoIs can be considered, such as actions on objects (e.g., a door that is being opened or closed). 
We assume that each FoI is in one of two different states, representing whether they are currently \emph{active} or \emph{idle}. To each FoI is associated an interval $[min, max]$, called the \emph{arity} of the FoI, representing how many persons minimum (resp. maximum) can activate this FoI at once. For instance, a motion sensor will be associated with the interval $[1, \infty]$ as it is activated if one persons enter its range and the activation remains the same whatever number of persons are present. A pressure sensor on a couch however could be associated with the interval $[1, 3]$ as one person at least needs to sit on the couch for the sensor to be activated but a maximum of 3 persons can sit together on the couch. By using FoIs instead of the actual sensor, we make it possible to represent different granularity of detection. It is possible for instance to estimate the number of persons sitting on a couch by analyzing the raw value given by the pressure sensor and thresholding it differently. This sensor could then be represented by several FoIs, each associated to a different threshold. We call \emph{activation line} the set of all states for all FoIs in the environment at a given point in time. 

We represent possible overlaps of FoIs in an undirected graph, called \emph{Co-activation Graph (CG)}.
Each node in the CG represents a FoI and two nodes are connected if a single person can activate both FoIs at the same time. A simple example would be a pressure sensor on a couch and a motion sensor in the room containing the couch. A single person could activate both FoIs by sitting on couch and being detected by the motion sensor.

Finally, we assume that the environment has only one \emph{entry point} from which a single person can enter or exit at a given time point, and that this entry point is monitored and therefore associated to a FoI. We also assume that the maximum number of persons that can be in the environment at once is bounded. We denote this number $\MaxNbPersons$.

\begin{figure}[htbp]
	\centering
	\includegraphics[width=0.8\linewidth]{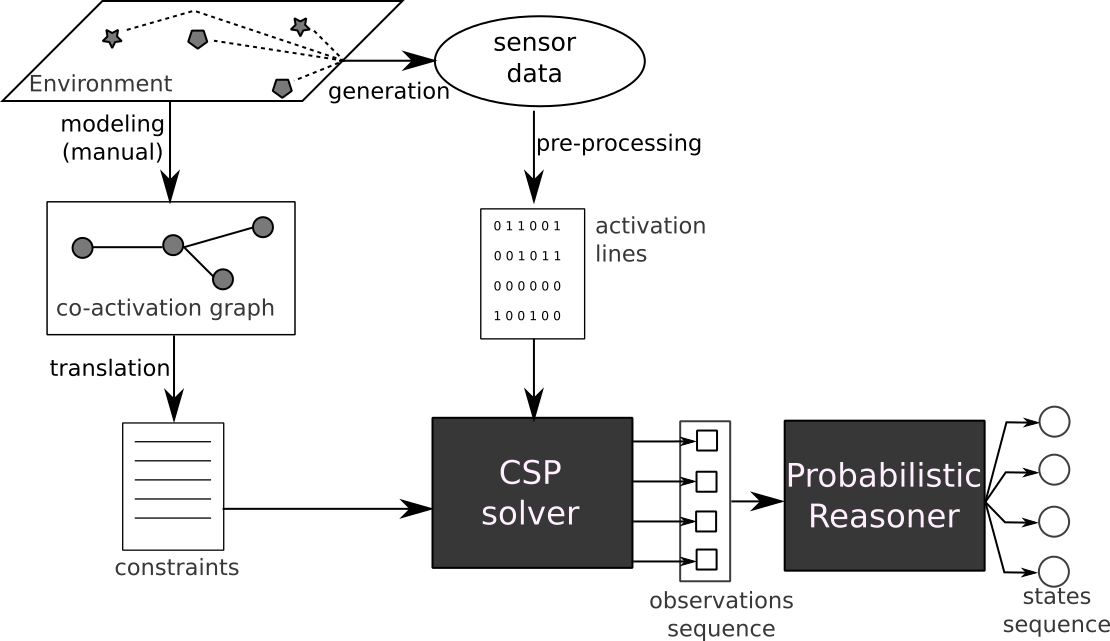}
	\caption{Overall diagram of our person counting system}
	\label{fig:overall}
\end{figure}

Figure \ref{fig:overall} illustrates the overall approach. The environment and sensors
need to be modeled to create the co-activation graph required
by the CSP solver (see below). 
The CSP solver also receives the current activation line representing the state of all FoIs. 
Using this input, the CSP produces an estimate of the number of people in the environment for each time point.
The Probabilistic Reasoner then uses these estimates as observations to generate a state sequence, in which each state contains the number of persons present in the environment at the associated time point. The reasoner uses a sliding window over the observations sequence in order to perform the counting online. 
We now describe this approach in detail over the next few sections.

\subsection{The Constraint Satisfaction Problem}

Constraint Satisfaction Problems (CSPs) \cite{dechter2003constraint} are modeled by providing a set of variables $X$, 
a set of domains $D[x]$ for each $x \in X$, and a set of constraints $C$ that specify allowed (or disallowed) relations between different variables.
A solution to a CSP is an assignment for each variable in $X$ that satisfies all constraints in $C$.

To get an estimate of the number of people in
an environment we create a CSP in the following way.
Let $P = \{ 1,\ldots,\MaxNbPersons\}$ represent a set of persons.
We create one variable $x \in X$ for each FoI
with $D[x] = 2^P$.
An assignment of a variable $x$ is
the set of persons observed by a specific FoI.
Further, as input for each FoI $x$ we assume
to know $\Activated_x \in \{ \True, \False \}$
(i.e., the activation line), as well as
$\Min_x \in \{1, \ldots, \MaxNbPersons\}$ and $\Max_x \in \{1, \ldots, \MaxNbPersons\}$
(the minimum and maximum number of persons indicated by
the FoI).
For each pair of FoIs $x_1$ and $x_2$ we use $\Linked(x_1,x_2) \in \{ \True, \False \}$
indicating whether or not two FoIs have an edge in the co-activation graph.

Given this information we impose the following constraints:
\begin{eqnarray}
 \forall x \in X &:& \neg \Activated_x \Leftrightarrow |x| = 0 \label{con1} \\
 \forall x \in X &:& \Activated_x \Leftrightarrow \Min_x \leq |x| \leq \Max_x \label{con2} \\
 \forall x_1,x_2 \in x &:& x_1 \cap x_2 = \emptyset \vee \Linked(x_1,x_2) \label{con3}
\end{eqnarray}

Constraint \ref{con1} states that if FoI $x$ is not activated it does not observe any person.
Constraint \ref{con2} states that if FoI $x$ is activated it observers any number of
persons between the minimum and maximum allowed for this FoI.
Finally, Constraint \ref{con3} states that two FoIs can only observe
the same person when they are neighbors in the co-activation graph.

To solve this problem we modeled it in the constraint processing language MiniZinc \footnote{http://www.minizinc.org/} and used the default solver that comes with the MiniZinc
software.  We chose constraint processing to solve this sub-problem as it is easy to model
and we can rely on existing solvers to create solutions relatively fast.
While we do not use minimization in order compute solutions 
fast enough, we assume that solutions to the CSP are close the minimal number of persons in the environment.
The reason for this is that the constraints above are much easier to satisfy when each FoI $x$ is assigned the smallest possible number of persons.

\subsection{The Probabilistic Reasoner}

The role of the probabilistic reasoner is to take a sequence of outputs produced by the CSP solver and refine them by considering \emph{aleatory} uncertainty and \emph{epistemic} uncertainty. Aleatory uncertainty is caused by randomness and variability in the system, in our case possible false sensor measures. Epistemic uncertainty is due to a lack of knowledge, in our case overlapping FoIs which can produce the same activation line for different numbers of persons within their range.  
To deal with both types of uncertainty, we need a reasoning model capable of: 
\begin{itemize}
	\item capturing and using a priori knowledge about the environment such as its dynamic, the relations between the environment state and the observed output, 
	\item reasoning sequentially over series of inputs. 
\end{itemize} 

Hidden Markov Models (HMM) are a very well studied class of Bayesian reasoning models capable of capturing such information. A HMM is made of hidden states that can generate observations. A transition matrix explicits the dynamic of the model, which respects the Markov Property, i.e., that the state at time $t$ only depends on the  state at time $t-1$. Finally, an emission matrix defines the probability with which a given state can generate a given observation. HMMs have been successfully used in various applications involving sequential reasoning, especially in speech and handwriting recognition~\cite{gales_application_2008,ploetz_markov_2009}. A complete overview of HMMs can be found in~\cite{rabiner_tutorial_1989}. 

In our framework, the variable being modeled in the HMM is the number of persons in the environment. Mathematically, our person counting HMM is described as follows: 
\begin{itemize}
	\item $S = \{s_0, \ldots , s_m\}$ is the set of states, with $ \forall s_i = [\NbPersons_i, \Activated_i]$ with $\NbPersons_i \in [0, \MaxNbPersons]$ being the number of persons in the environment and $\Activated_i \in \{\True, \False\}$ is the state of the entry point. By construction, there is $|S| = (\MaxNbPersons+1) * 2$ states in the state space.  
	\item $V = \{v_0, \dot , v_n\}$ is the observation alphabet, with $\forall v_i = [\CspEstimation_i, \Activated_i]$ with $\CspEstimation_i \in [0, \MaxNbPersons]$ being the estimate given by the CSP solver and $\Activated_i \in \{\True, \False\}$ being the same state of the entry point. By construction, there are $|V| = (\MaxNbPersons+1) * 2$ elements in the observation alphabet. 
	\item $Q = q_1, \ldots, q_T$ is a fixed sequence of $T$ states 
    \item $O = o_1, \ldots, o_T$ is a fixed  sequence of $T$ observations
	\item $\pi = [\pi_i], \pi_i = P(q_1 = s_i)$ is the initial probability array. Without prior information, $\pi$ is a uniform distribution. 
	\item $A = [a_{ij}], a_{ij} = P(q_t = s_j | q_{t-1} = s_i)$ is the transition matrix, storing the probability of state $s_j$ following $s_i$. 
	\item $B = [b_i(k)], b_i(k) = P(o_t = v_k | q_t = s_i)$ is the observation matrix, storing the probability that observation $v_{k}$ is produced from state $s_i$. 
\end{itemize}

Note that the states in our HMM do not only include hidden variables but also part of the observation. The reason for this design choice is due to the fact that considering the state of the entry point as part of the observation enables us to model the dynamic of the system with a finer granularity than considering only the result of the CSP. Indeed, this FoI has the specificity to announce when somebody is likely to have entered or to left the environment. However, by considering this FoI in the observation, our system looses its Markovian property as the state at time $t$ depends on both the state and the observation at time $t-1$. By integrating the FoI in the state as well, we can restore the Markovian property. 

The transition matrix is defined based on static, likely and unlikely transitions. Static transitions are simply transitions from $s_i$ to $s_i$. Definition \ref{def:likely} gives the likely transitions for normal case, i.e. $\omega_i  \neq 0$ and $\omega_i \neq \Omega$ (Equation ~\ref{eqn:likely_normal}), and for the border cases (Equation~\ref{eqn:likely_extreme}). 
\begin{definition}[Likely transition]
\label{def:likely}
	A transition between $s_i = [\NbPersons_i, \Activated_i]$ and $s_j = [\NbPersons_j, \Activated_j]$ is considered likely if and only if for $0 < \NbPersons_i < \MaxNbPersons$: 
	\begin{equation}
	\label{eqn:likely_normal}
	\left\{
	\begin{array}{@{}lr@{}}
		\begin{array}{r@{}}
		\NbPersons_j = \NbPersons_i \text{ and } \Activated_j = \True \\
		\text{or } \NbPersons_j = \NbPersons_i + 1 \text{ and } \Activated_j = \True\\
		\end{array} & 
		\for \Activated_i = \False \\ \\
		\begin{array}{r@{}}
		\NbPersons_j = \NbPersons_i \text{ and } \Activated_j = \False \\
		\text{or } \NbPersons_j = \NbPersons_i-1 \text{ and } \Activated_j = \False\\
		\end{array} & 
		\for \Activated_i = \True \\
	\end{array} \right.
	\end{equation}
	
	The following transitions are also considered likely: 
	\begin{equation}
	\label{eqn:likely_extreme}
	\begin{array}{@{}lr@{}}
		s_i = [0,\False] \text{ to } s_j = [1,\True] \\
		s_i = [0,\True] \text{ to } s_j = [0,\False] \\
		s_i = [0,\True] \text{ to } s_j = [1,\True] \\
		s_i = [\MaxNbPersons, \False] \text{ to } s_j = [\MaxNbPersons, \False] \\
		s_i = [\MaxNbPersons, \True] \text{ to } s_j = [\MaxNbPersons - 1, \True]
	\end{array}
	\end{equation}
\end{definition}
The intuition behind this modeling is that if the entry point becomes activated ($\Activated_i = \False$ and $\Activated_j = \True$), it is likely that either someone already in the environment activated it or someone just entered the environment. On the opposite, if the entry point was activated and is now idle ($\Activated_i=\True$ and $\Activated_j=\False$), it is likely that either someone stayed in the environment but left the FoI's range or someone just left the environment. By construction there is then 2 likely transitions for the normal cases and 1 or 2 for the special cases. 

All transitions that are neither static nor likely are considered unlikely. 
From the previous definition, we can derive the transition probabilities for  static and likely transitions. All static transitions have a probability $a_{ii} = p_{static}$. The probabilities of likely transitions are computed as follows for the special cases: 
\begin{equation}
\label{eqn:transition_extreme}
\begin{array}{@{}lr@{}r@{}}
	a_{ij} = p_{likely} & \for & s_i = [0,0] \text{ and } s_j = [1,1] \\
	a_{ij} = \frac{p_{likely}}{2} & \for & s_i = [0,1] \text{ and } s_j = [0,0] \\
	a_{ij} = \frac{p_{likely}}{2} & \for &  s_i = [0,1] \text{ and }s_j = [1,1] \\
	a_{ij} = p_{likely} & \for & s_i = [\MaxNbPersons, 0] \text{ and }s_j = [\MaxNbPersons, 1] \\
	a_{ij} = p_{likely} & \for & s_i = [\MaxNbPersons, 1] \text{ and }s_j = [\MaxNbPersons - 1, 0]
\end{array}
\end{equation}
For the normal cases: 
\begin{equation}
\label{eqn:transition normal}
	\begin{array}{@{}l}
		a_{ij} = \frac{p_{likely}}{2}
	\end{array}
\end{equation}

The probabilities for the unlikely transitions can be easily derived from equations \ref{eqn:transition_extreme} and \ref{eqn:transition normal} : 
\begin{equation}
	a_{ij} = \frac{p_{unlikely}}{|S|^2 - (n_{i, likely} + 1)}
\end{equation} 
where $n_{i, likely}$ is the number of likely transitions from state $s_i$. 

The emission matrix is defined similarly by considering impossible, correct, probable and unprobable emissions. It also uses the fact that, as discussed previously, 
the CSP is more likely to underestimate the number of persons in the environment rather than overestimate it.  Given this knowledge and considering a state $s_i = [\NbPersons_i, \Activated_i]$ and an observation $v_k = [\CspEstimation_k, \Activated_k]$, we know that: 
\begin{itemize}
	\item the emission of $v_k$ in state $s_i$ is impossible by construction iff $\Activated_i \neq \Activated_k$, 
	\item the emission of $v_k$ by $s_i$ is correct iff $\NbPersons_i = \CspEstimation_i$ and $\Activated_i = \Activated_k$, 
	\item the emission of $v_k$ while in $s_i$ is probable iff $\NbPersons_i > \CspEstimation_k$ and $\Activated_i = \Activated_k$, 
	\item the emission of $v_k$ while in $s_i$ is unprobable iff $\NbPersons_i < \CspEstimation_k$ and $\Activated_i = \Activated_k$. 
\end{itemize}
From this it is possible to derive the emission probabilities: 
\begin{equation}
	\label{eqn:emission_normal}
	\begin{array}{@{}lr@{}r@{}}
		b_i(k) = p_{correct} & \for & \CspEstimation_k = \NbPersons_i \\
		b_i(k) = \frac{p_{probable}}{\NbPersons_i} & \for & \CspEstimation_k < \NbPersons_i \\
		b_i(k) = \frac{p_{unprobable}}{\MaxNbPersons - \CspEstimation_k} & \for &  \CspEstimation_k > \NbPersons_i \\
	\end{array}
\end{equation}

%% file: sections/3-evaluation.tex
\subsection{Evaluation Design}
\label{sec:evaluation-design}

\begin{figure}[htbp!]
	\centering
	\begin{subfigure}{0.75\textwidth}
		\includegraphics[width=\textwidth]{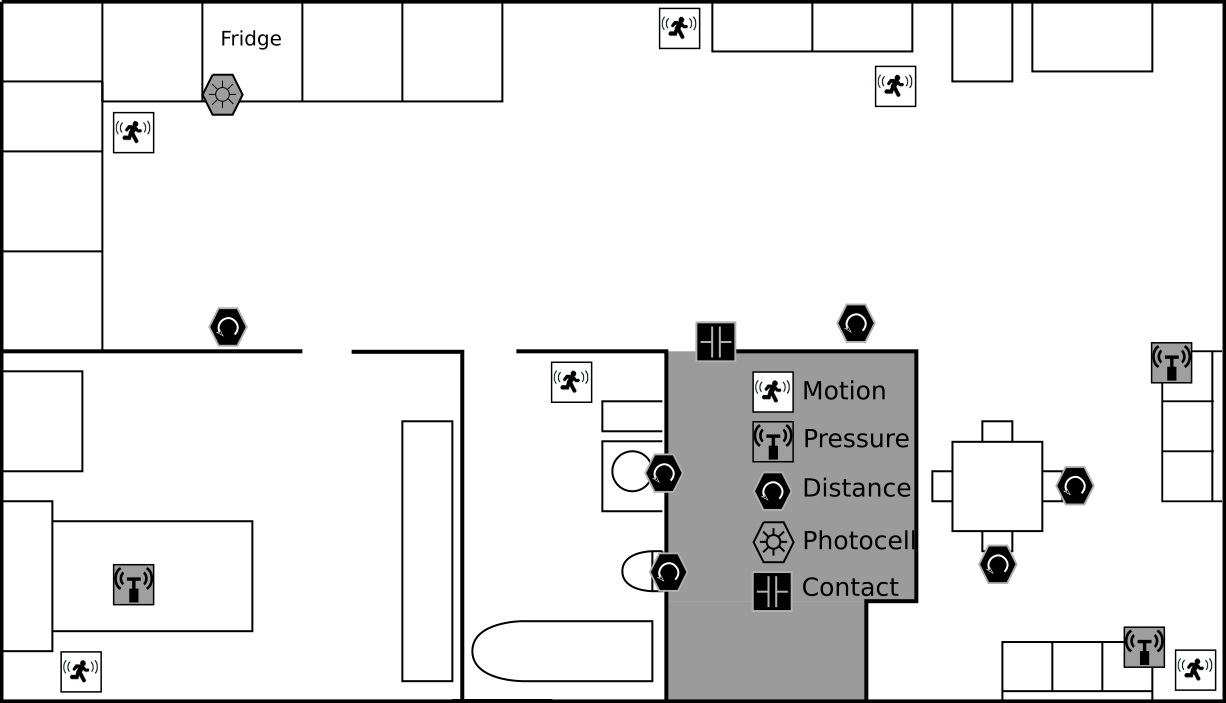}
		\subcaption{Map of ARAS (without motion sensors) and ARAS-FC environments. ARAS-FC is the full-coverage version of ARAS, in which motion sensors have been added in each rooms. Figure modified from \cite{alemdar_aras_2013}.}
		\label{fig:aras_map}
        \end{subfigure}
              
	\begin{subfigure}{0.75\textwidth}
		\includegraphics[width=\textwidth]{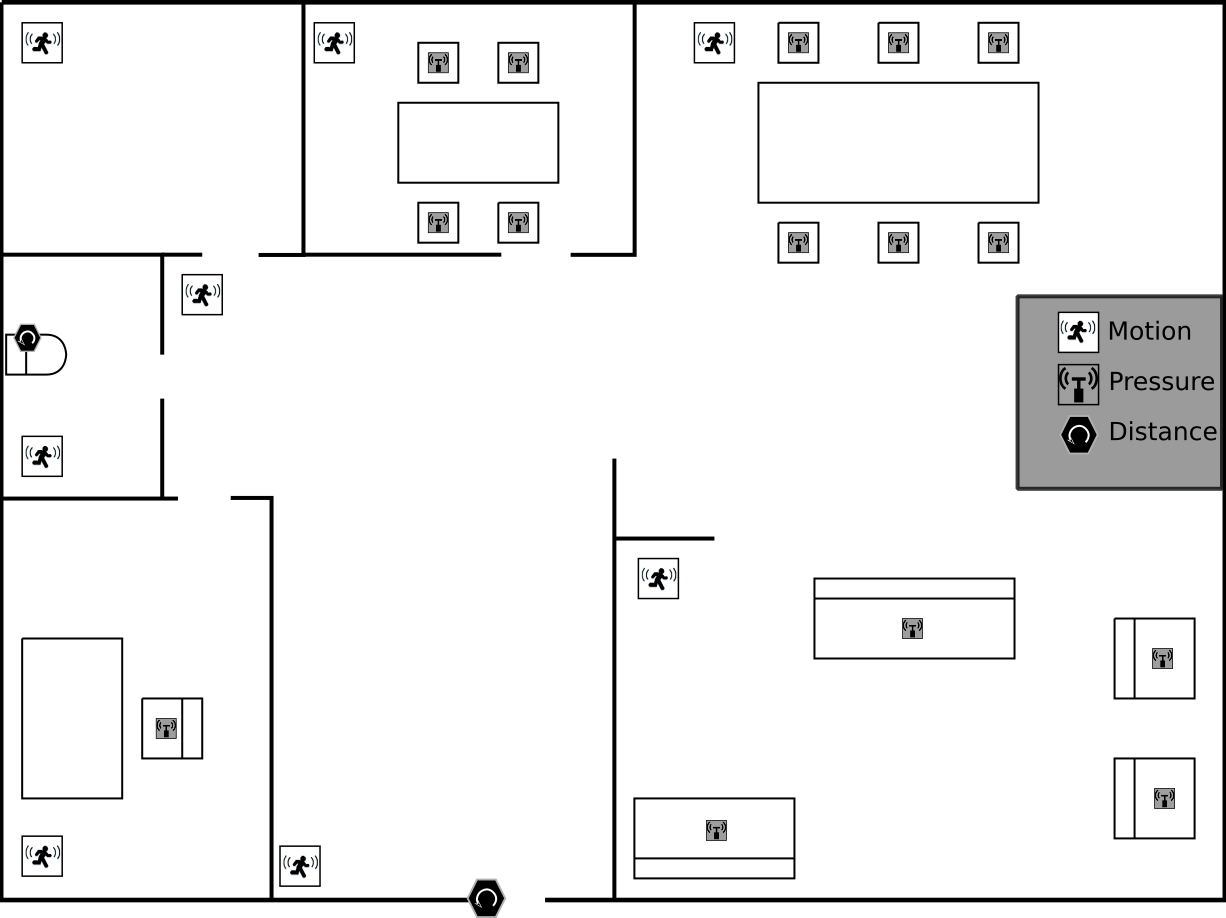}
		\subcaption{Map of the House 2 environment with sensors}
		\label{fig:home2_map}
	\end{subfigure}
	\caption{Map of test environments with sensors}
\end{figure}

To evaluate our system, we used simulated and real datasets. Simulated data was necessary for two reasons. First available datasets only consider one or two occupants while it is important to evaluate the performance of our system on larger number of occupants. Second, available datasets have limitations in the size of the environment they consider (usually small) and its coverage (usually partial), while we wished to evaluate the difference in the performance of our system depending on the size and the coverage. For these reasons, we use three different configurations: 
\begin{enumerate}
\item ARAS: small environment, partial coverage, reproducing the House A from \cite{alemdar_aras_2013}, presented on Figure \ref{fig:aras_map}. Tests have been performed on simulated and real data for this environment. 
\item ARAS-FC: small environment, full coverage, modified from ARAS by adding motion sensors in each room to create a full coverage, presented on Figure \ref{fig:aras_map}. Due to the lack of available dataset for this environment (the ARAS dataset considering only partial coverage), tests have been performed on simulated data only. 
\item House 2: large environment, full coverage, presented on Figure \ref{fig:home2_map}. This environment has been chosen to study the different in performance between a small and a large environment. Due to the lack of available dataset for this environment, tests have been performed on simulated data only. 
\end{enumerate}

For the three environment, one FoI has been created for each sensor present.

The real dataset used for the ARAS environment is the dataset provided in \cite{alemdar_aras_2013}, which consists of 30 days of data with one activation line per second, i.e. 86400 activation lines per day. The first 3 days of data have been used to learn the parameters of the system while the 27 remaining days have been used to evaluate the system. 

Data were simulated using  a pseudo-random walk in a Markov Chain. A node in the Markov Chain represents a possible Feature of Interest. One node represents the outside of the apartment. Two nodes are connected if the user can access the ending FoI from the starting FoI without activating any other FoI. The transitions probabilities were decided arbitrarily to represent plausible transitions. Each node is also associated with a maximum number of persons that can be present on this node at the same time. Multi-agent pseudo random walks are performed over the Markov Chain to generate datasets: at each time step, all the agents randomly select an accessible node with available space and move to this location. Data simulated this way is obviously not realistic, i.e., the agents are going to perform a sequence of movements without following any specific activity plan, changing location and therefore activated node much more often than a real person would. However, despite this lack of realism, simulating data this way allowed us to obtain relatively large datasets quickly on various number of persons. A total of 1000 activation lines have been generated and used to learn the parameters of the HMMs and 10000 have been used to evaluate the system, distributed in 10 runs of 1000 lines. 

For each environment we optimized 9 different HMMs, one per maximum number of persons  allowed ranging from 2 to 10. Due to a lack of space, only the results for the HMMs created with 4, 6, 8 and 10 persons maximum (referred respectively as HMM-4, HMM-6, HMM-8 and HMM-10) will be presented in this paper. The other HMMs showed similar trends. The optimization of the HMM has been done by learning the transition and the observation matrices. The learning of the transition matrix (resp. observation matrix) has been performed by calculating the frequency of likely and unlikely transitions (resp. correct, probable and unprobable emissions). For tests using simulated data, the learning has been performed over 1000 activation lines, different from the ones used for testing. For tests using real data, the learning has been performed over 3 days of data, i.e. $3\times86400$ activation lines and the been performed over the 27 remaining days.

In order to assess the usefulness of the HMM, we perform each experiment using only the CSP or the combination CSP+HMM. This step has been done to ensure that the HMM was indeed useful and that the CSP alone was not enough to achieve acceptable prediction. Finally, preliminary experiments not presented in this paper for lack of space showed that the size of the sliding window, when 10 or above, do not impact significantly the results of the prediction. Therefore, these experiments have been conducted with a sliding window of size 10. 

\subsection{Results}
\subsubsection{Simulated data}
\begin{figure}[htbp]
	\centering
	\begin{subfigure}[b]{0.40\textwidth}
		\includegraphics[width=\textwidth]{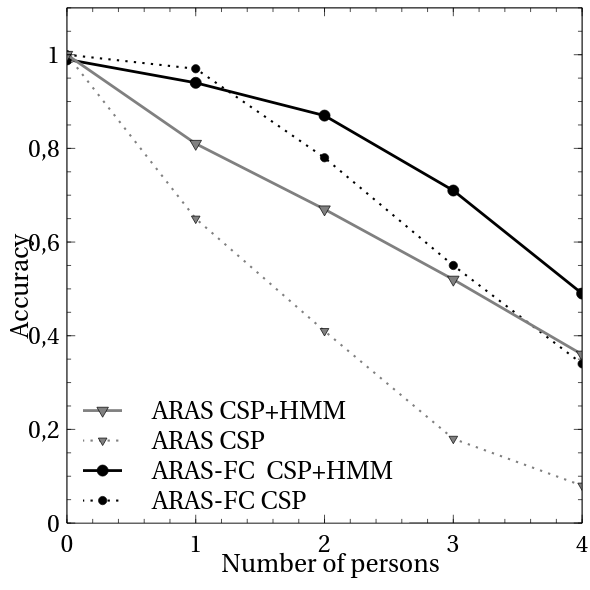}
		\caption{HMM-4}
		\label{fig:4_persons}
	\end{subfigure}
	 \begin{subfigure}[b]{0.40\textwidth}
	 	\includegraphics[width=\textwidth]{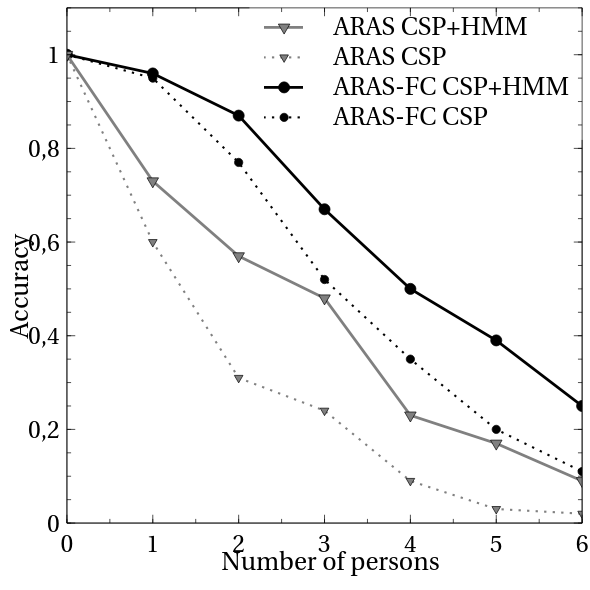}
	 	\caption{HMM-6}
	 	\label{fig:6_persons}
	 \end{subfigure}
	 \begin{subfigure}[b]{0.40\textwidth}
	 	\includegraphics[width=\textwidth]{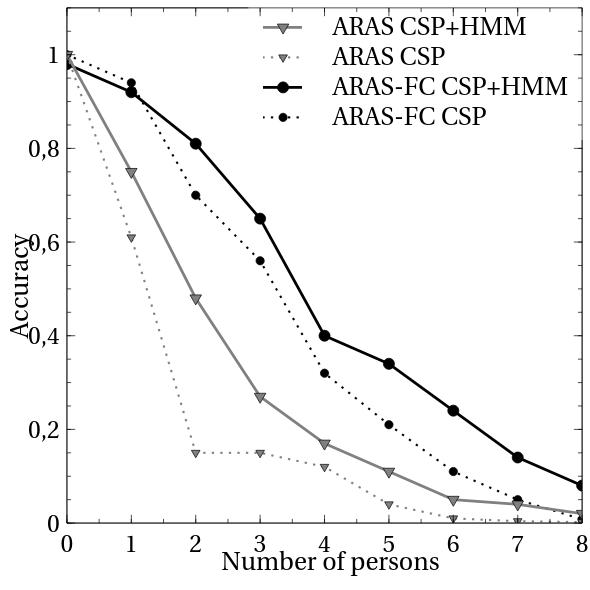}
	 	\caption{HMM-8}
	 	\label{fig:8_persons}
	 \end{subfigure}
	\begin{subfigure}[b]{0.40\textwidth}
		\includegraphics[width=\textwidth]{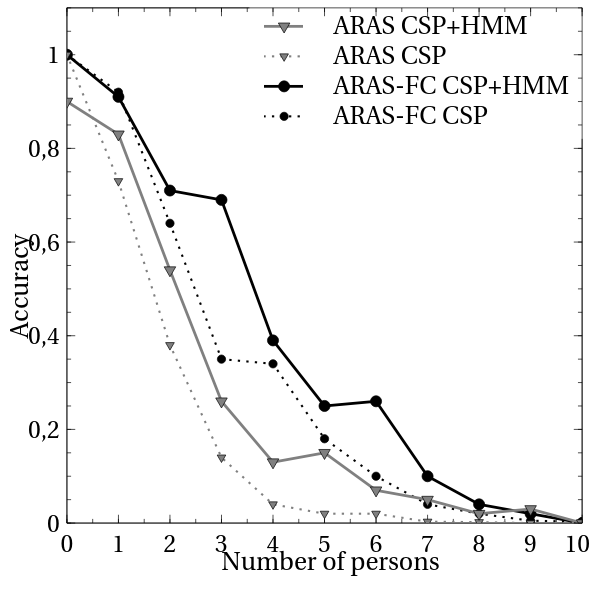}
		\caption{HMM-10}
		\label{fig:10_persons}
	\end{subfigure}
	\caption{Accuracy obtained on ARAS and ARAS-FC environments}\label{fig:aras}
      \end{figure}

Figure \ref{fig:aras} presents the accuracy obtained for the ARAS and ARAS-FC environments.  Accuracy is defined as the proportion of exact matches between the predicted and actual number.

\begin{figure}[htbp]
	\centering
	\begin{subfigure}[b]{0.40\textwidth}
		\includegraphics[width=\textwidth]{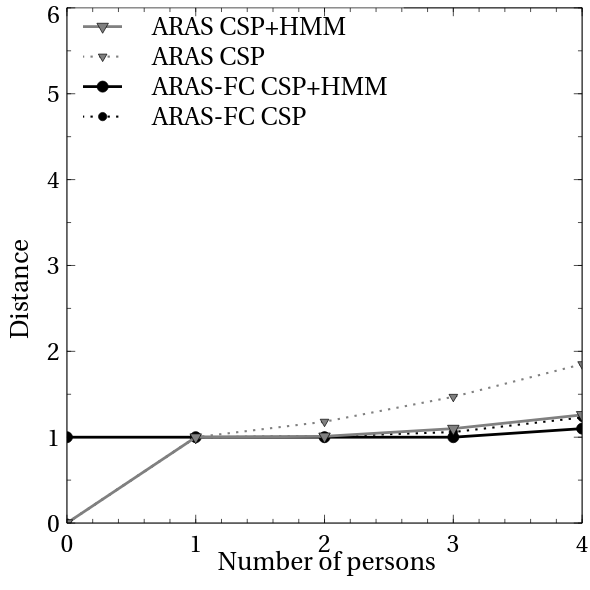}
		\caption{HMM-4}
		\label{fig:dist_aras_4_persons}
	\end{subfigure}
	 \begin{subfigure}[b]{0.40\textwidth}
	 	\includegraphics[width=\textwidth]{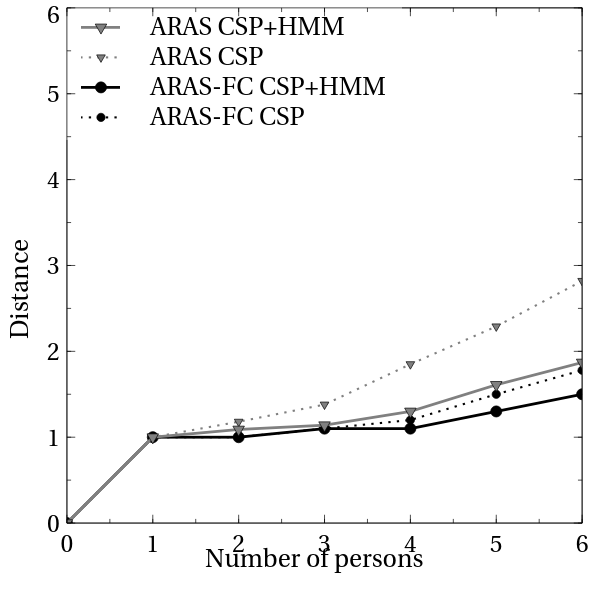}
	 	\caption{HMM-6}
	 	\label{fig:dist_aras_6_persons}
	 \end{subfigure}
	 \begin{subfigure}[b]{0.40\textwidth}
	 	\includegraphics[width=\textwidth]{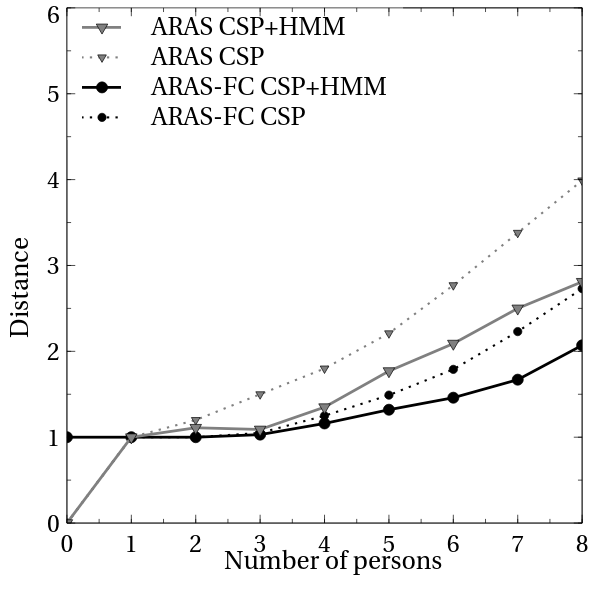}
	 	\caption{HMM-8}
	 	\label{fig:dist_aras_8_persons}
	 \end{subfigure}
	\begin{subfigure}[b]{0.40\textwidth}
		\includegraphics[width=\textwidth]{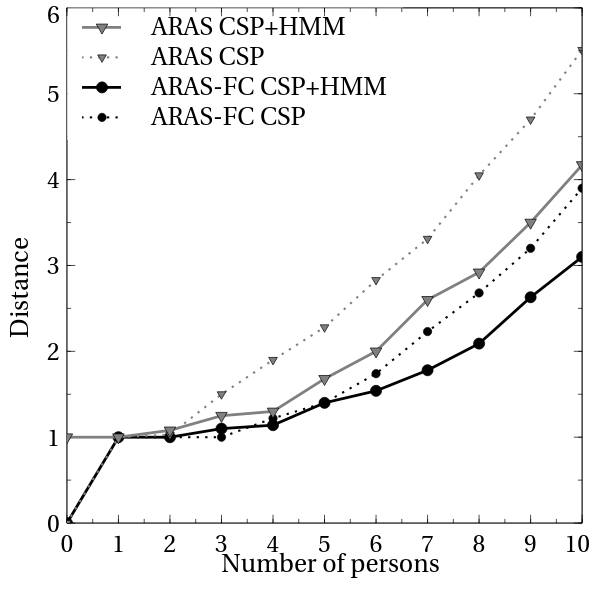}
		\caption{HMM-10}
		\label{fig:dist_aras_10_persons}
	\end{subfigure}
	\caption{Distance obtained on ARAS and ARAS-FC environments}
      \label{fig:distance-aras}
\end{figure}

Figure \ref{fig:distance-aras} presents the average distance between the predicted number of persons and the actual number of persons for the different ARAS environment. Note that the distance has been only computed when the prediction was incorrect. 
From the experiments, we observed first of all that the accuracy of the system in the ARAS-FC environment is much better than the one in ARAS with an improvement of 46\% on average
. 
This highlights the importance of choosing and positioning properly the sensors when performing counting with pervasive sensors. The second point to be noted is that, as expected, the system with HMM is performing better than the system without HMM with an average improvement of 38\% for ARAS and 17\% for ARAS-FC 
. Third, we noticed that the performance of the system decreases slightly when the HMM is optimized for a higher maximum number of persons. 
Finally, we noticed that the accuracy of the system is globally good for 0 to 3 persons (99\% to 68\% accuracy on average for ARAS-FC) but decrease dramatically for 4 persons and more. (45\% to close to 0\% on average). Our hypothesis to explain this difference is twofold: 
\begin{enumerate}
	\item the size of the environment (rather small) makes it impossible to discriminate between persons after a certain number. Indeed, the number of FoIs that a person can activate and have a finite arity is rather small and therefore limits the possibility of detection. This hypothesis will be enforced by our experiments in the House 2. 
	\item the behavior of the simulated agents are very erratic and they change location during the simulation more often and with less coherence than real agents, which creates more difficult situations for the reasoner. Indeed, HMMs in general are used and capable of recognizing patterns and changes in these patterns by making use of sequences of observations. In our case, there is no pattern not make use of and the changes in the environment are very erratic, making it more difficult to predict. We could hypothesize that a real agent would follow more consistent patterns that the Markov Chains we used for simulated data failed to capture. More experiments either with real or more accurate simulated data would be needed to test this hypothesis. 
\end{enumerate}
However, it is interesting to note that despite the drop in accuracy, the distance between the predicted number of persons and the actual one remains low in the case of ARAS-FC CSP+HMM. 

\begin{figure}[htbp]
	\centering
	\begin{subfigure}[b]{0.40\textwidth}
		\includegraphics[width=\textwidth]{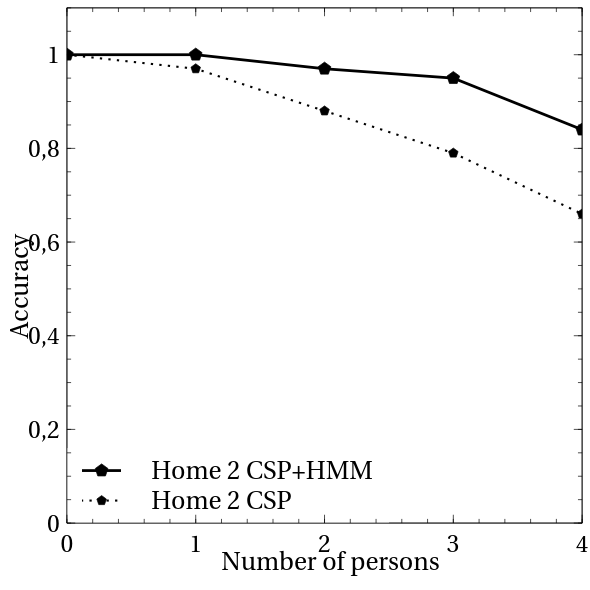}
		\caption{HMM-4}
		\label{fig:home2_4_persons}
	\end{subfigure}
	 \begin{subfigure}[b]{0.40\textwidth}
	 	\includegraphics[width=\textwidth]{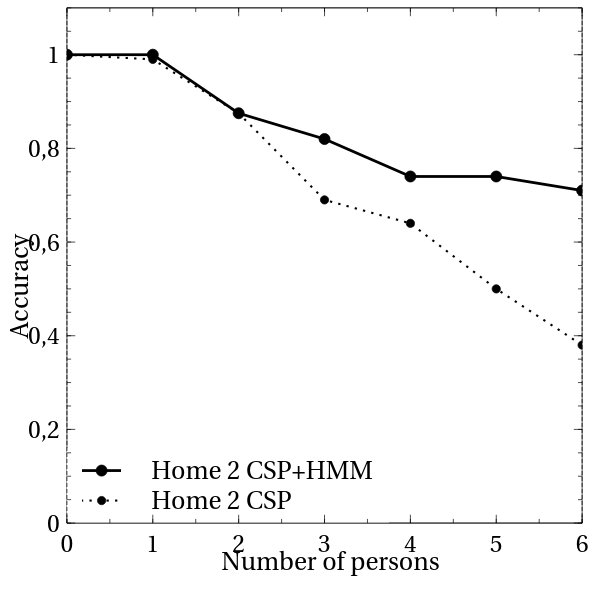}
	 	\caption{HMM-6}
	 	\label{fig:home2_6_persons}
	 \end{subfigure}
	 \begin{subfigure}[b]{0.40\textwidth}
		\includegraphics[width=\textwidth]{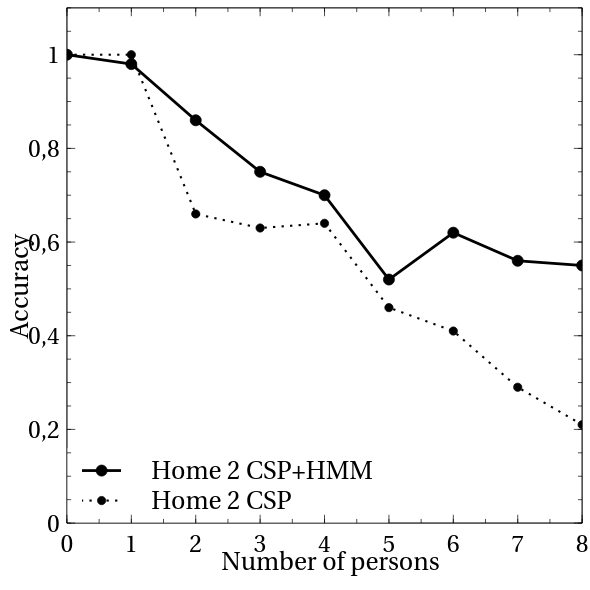}
	 	\caption{HMM-8}
	 	\label{fig:home2_8_persons}
	 \end{subfigure}
	\begin{subfigure}[b]{0.40\textwidth}
		\includegraphics[width=\textwidth]{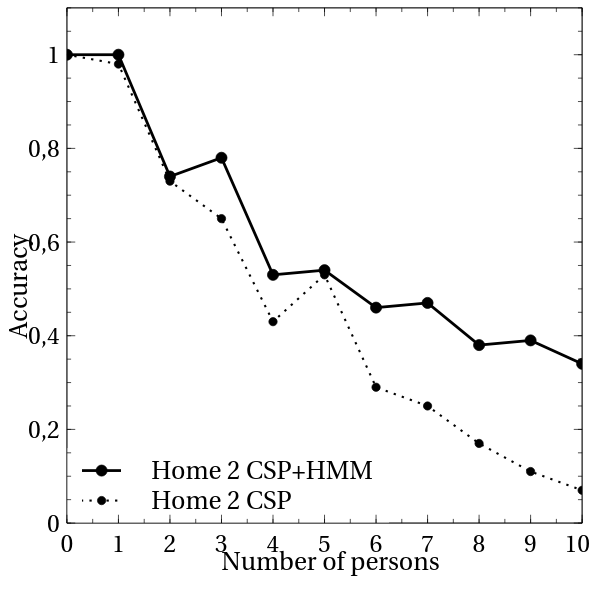}
		\caption{HMM-10}
		\label{fig:home2_10_persons}
	\end{subfigure}
	\caption{Accuracy obtained on the House 2 environment}
	\label{fig:home2}
\end{figure}

Figure \ref{fig:home2} presents the detection accuracy obtained for the Home 2, with and without the HMM and for various maximum number of persons. 
As before, we noticed that in this environment the system is performing better with HMM than without (improvement of 19\%
). We also noticed that the overall detection accuracy is much better in Home 2 than in ARAS-FC for larger number of persons, which tends to confirm the hypothesis that the size of the environment (and more specifically the number of FoIs associated to finite arity) affects the performance of our system.

\begin{figure}[htbp]
	\centering
	\begin{subfigure}[b]{0.40\textwidth}
		\includegraphics[width=\textwidth]{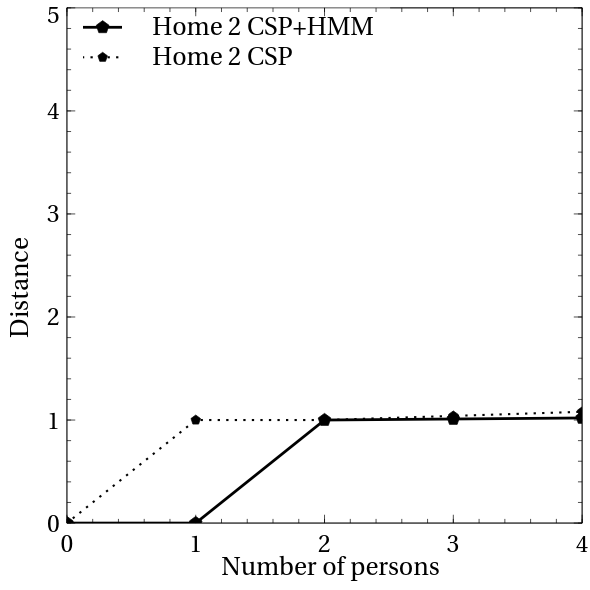}
		\caption{HMM-4}
		\label{fig:dist_home2_4_persons}
	\end{subfigure}
	 \begin{subfigure}[b]{0.40\textwidth}
	 	\includegraphics[width=\textwidth]{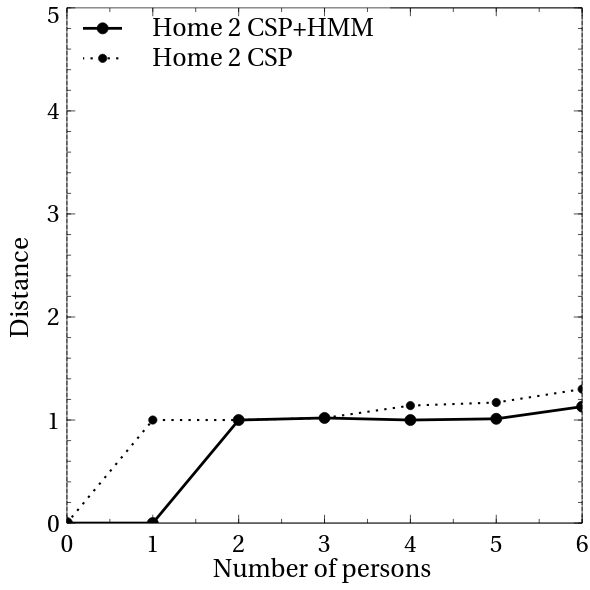}
	 	\caption{HMM-6}
	 	\label{fig:dist_home2_6_persons}
	 \end{subfigure}
	 \begin{subfigure}[b]{0.40\textwidth}
		\includegraphics[width=\textwidth]{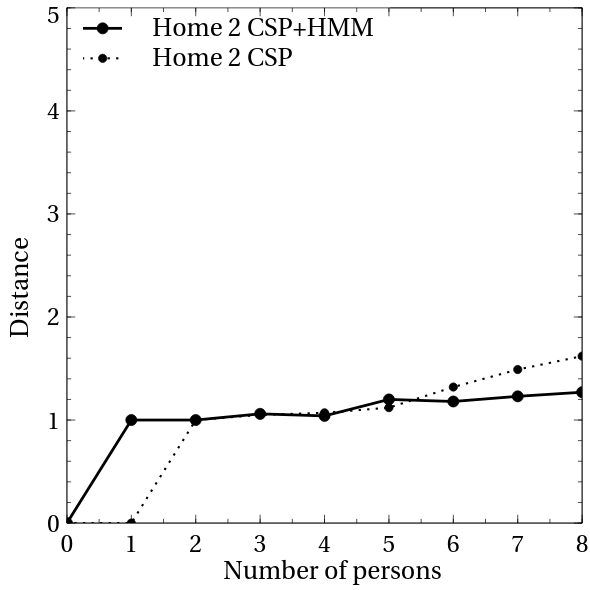}
	 	\caption{HMM-8}
	 	\label{fig:dist_home2_8_persons}
	 \end{subfigure}
	\begin{subfigure}[b]{0.40\textwidth}
		\includegraphics[width=\textwidth]{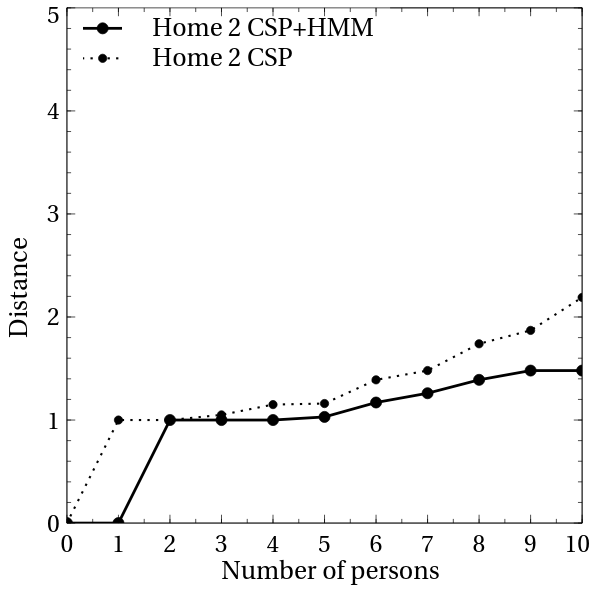}
		\caption{HMM-10}
		\label{fig:dist_home2_10_persons}
	\end{subfigure}
	\caption{Distance obtained on the House 2 environment}
	\label{fig:distance-home2}
\end{figure}

Figure \ref{fig:distance-home2} presents the average distance between the predicted number of persons and the actual number of persons, restricted to the cases where the prediction was incorrect. 
Once again, we notice that despite the drop in the accuracy, the distance between predicted and actual number of persons remains rather low, even for large number of persons.


\subsubsection{Real data}
\label{subsec:real}
Tests on real data have been performed on the ARAS environment using both HMM-4 and HMM-10 in order to compare the results. With both HMM-4 and HMM-10, our system reached an overall detection accuracy of 80\% (against 83\% for HMM-4 and 76\% for HMM-10 with simulated data).
The differences between the results obtained with simulated and real data are not significant enough to enable us to confirm or contradict our hypothesis that real data are easier to reason upon than simulated data. More experiments, especially with higher number of persons, are required but the lack of such data currently makes it impossible to do so. However, the high percentage of accurate detection with real data is by itself a promising result regarding the ability of our system to detect accurately the number of persons in the environment with real data. This result, once more, needs to be confirmed with other datasets considering different number of persons.


%% file: sections/4-discussion.tex

We presented a system capable of counting the number of persons present in an environment using pervasive sensors and a combination of a CSP solver and a Hidden Markov Model. Our system is capable of performing online counting with an overall accuracy of 80\% on both simulated and real data. These results have been achieved while learning the parameters of the HMM using very basic statistic computation.  

Extensive experiments enabled us to identify a set of factors that influence the result of this system. First of all, the coverage of the environment by the sensors is of high importance. For good results, the environment needs to be as fully covered by sensors as possible, meaning that the number of places in which a person can be located without being sensed by any sensor needs to be as small as possible. Then, the maximum number of persons allowed in the environment must be realistic enough to capture various situations but not too high to keep the detection accurate. Even with a higher maximum number of persons allowed than required, the system is still performing very well to discriminate between 0, 1, 2 or more occupants, making it reliable to detect guests in various E-Health and Ambient Assisted Living systems.

Our system is based on different assumptions that could be relaxed in future work. First of all, it assumes a single entry point in the environment. While this assumption holds for a lot of environments such as regular apartments, other environments such as houses might present additional entry points (e.g., a garage door, or a back-door to a garden). Our system needs to be adapted for these situations. Then, the second and most constraining hypothesis is the maximum number of persons considered. This assumption had been made to avoid dealing with infinite sets of possible states, which is a common difficulty in reasoning. However, this constrains the user to choose an appropriate maximum since it impacts the performances of our system. Relaxing this assumption while keeping a good detection accuracy is an important step to take. 

Apart from relaxing these assumptions, we would like to investigate ways to increase the accuracy of the system.
One possibility in this direction, could be to model the environment in a more fine-grained manner (e.g., per room instead of per home). There are several ways to attempt improving the counting performance when a large number of persons are involved, such as providing  ground truth to the system or a more complex formulation of the HMM, especially considering likely and unlikely transitions. In addition  Finally, it would of interest to establish a theoretical baseline of how well any counting system can perform with a given set of sensors.


